\documentclass[letterpaper, 10 pt, conference]{ieeeconf}  

\IEEEoverridecommandlockouts                              

\usepackage{arydshln}
\usepackage{graphicx}
\usepackage{amsmath}
\usepackage{amssymb}
\usepackage{booktabs}
\usepackage{cite}
\usepackage[dvipsnames]{xcolor}
\usepackage{multirow}
\usepackage{hhline}
\usepackage{hyperref}
\hypersetup{
    colorlinks=true,
    linkcolor=black,
    filecolor=magenta,      
    urlcolor=black}
\usepackage[capitalize]{cleveref}
\usepackage{subcaption}
\title{\LARGE \bf
MS3D: Leveraging Multiple Detectors for Unsupervised Domain Adaptation in 3D Object Detection
}

\author{Darren Tsai, Julie Stephany Berrio, Mao Shan, Eduardo Nebot and Stewart Worrall%
\thanks{This work has been supported by the Australian Centre for Field Robotics (ACFR) and ARC LIEF grant LE200100049 Whopping Volta GPU Cluster - Transforming Artificial Intelligence Research. (Corresponding author: Darren Tsai.)}
\thanks{The authors are with the Australian Centre for Field Robotics (ACFR) at the University of Sydney (NSW, Australia). E-mails: {\small{\{d.tsai, j.berrio, m.shan, e.nebot, s.worrall}\}@acfr.usyd.edu.au}}%
}

\begin{document}

\maketitle
\thispagestyle{empty}
\pagestyle{empty}

\begin{abstract}

We introduce Multi-Source 3D (MS3D), a new self-training pipeline for unsupervised domain adaptation in 3D object detection. Despite the remarkable accuracy of 3D detectors, they often overfit to specific domain biases, leading to suboptimal performance in various sensor setups and environments. Existing methods typically focus on adapting a single detector to the target domain, overlooking the fact that different detectors possess distinct expertise on different unseen domains. MS3D leverages this by combining different pre-trained detectors from multiple source domains and incorporating temporal information to produce high-quality pseudo-labels for fine-tuning. Our proposed Kernel-Density Estimation (KDE) Box Fusion method fuses box proposals from multiple domains to obtain pseudo-labels that surpass the performance of the best source domain detectors. MS3D exhibits greater robustness to domain shift and produces accurate pseudo-labels over greater distances, making it well-suited for high-to-low beam domain adaptation and vice versa. Our method achieved state-of-the-art performance on all evaluated datasets, and we demonstrate that the pre-trained detector's source dataset has minimal impact on the fine-tuned result, making MS3D suitable for real-world applications. Our code is available at \url{https://github.com/darrenjkt/MS3D}.

\end{abstract}

\section{Introduction}
3D object detection is a vital component of AVs, serving as a fundamental building block for critical downstream tasks such as path prediction. The field has benefited greatly from the availability of large-scale AV datasets \cite{geiger2012kitti, sun2020waymo, caesar2020nuscenes, woven2019lyft}, which have fueled research in the development of 3D detectors. Unfortunately, directly applying these 3D detectors to new datasets results in suboptimal performance due to domain shift. Factors such as variations in lidars, geographic regions, scenarios, and sensor setups pose significant challenges for the effective testing and deployment of 3D object detectors. Traditionally, the approach to improving detectors for new datasets is to manually annotate the data for fine-tuning the existing detector. However, this approach is both costly and labour-intensive, therefore highlighting the need for effective approaches to adapt 3D detectors trained on labelled source domains to new, unlabelled target domains. This task is called unsupervised domain adaptation (UDA).

\begin{figure}[t]
  \centering
  \includegraphics[width=0.96\linewidth]{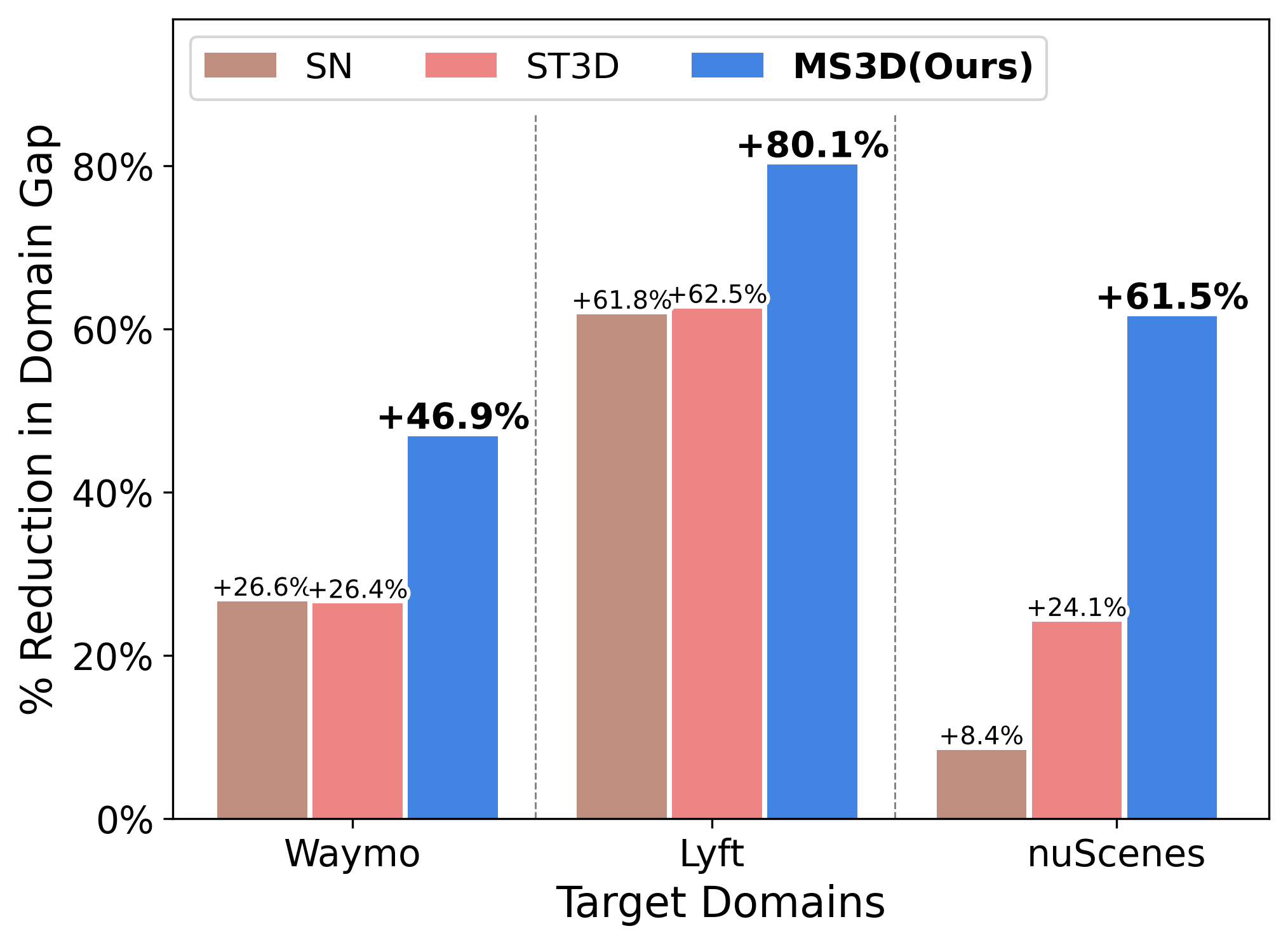}
  \caption{Performance of MS3D compared to SN \cite{wang2020train} and ST3D \cite{yang2021st3d} for Lyft $\rightarrow$ Waymo, Waymo $\rightarrow$ Lyft and Waymo $\rightarrow$ nuScenes (from left to right). Graph shows how much these methods have reduced the domain gap (in $\text{AP}_{\text{3D}}$ at IoU=0.7) between source-only (no UDA, i.e., 0\%), and fine-tuned with target domain labels (i.e., 100\%).}
  \label{fig:closed_gap_comparison}
  \vspace{-6mm}
\end{figure}

In this paper, we introduce Multi-Source 3D (MS3D), a novel pipeline for generating high-quality pseudo-labels on an unlabelled dataset for self-training in the context of UDA for 3D Object Detection. Unlike existing methods, MS3D takes advantage of the collective knowledge of multiple detectors to generate pseudo-labels on a single novel dataset. Moreover, by accumulating frames, MS3D improves the robustness of detection boxes on static objects.


One of the key benefits of MS3D is its ability to adapt robustly to a wide range of domain shifts by leveraging the expertise of multiple detectors. When using self-training in the standard single source-target domain adaptation setting, the choice of source dataset and detector can significantly impact fine-tuning performance. In practice, it is challenging to evaluate and identify the optimal source dataset and detector on new target datasets due to lack of labelled data. However, with MS3D, we eliminate the need to select a single pre-trained detector for self-training. We demonstrate that the combined pseudo-labels of multiple detectors are of higher quality than those of the optimal individual source-trained detector. As a result, our fine-tuned performance outperforms state-of-the-art single-source UDA methods as shown in \cref{fig:closed_gap_comparison}.



\begin{figure*}
  \vspace{2mm}
  \centering
  \includegraphics[width=0.99\linewidth]{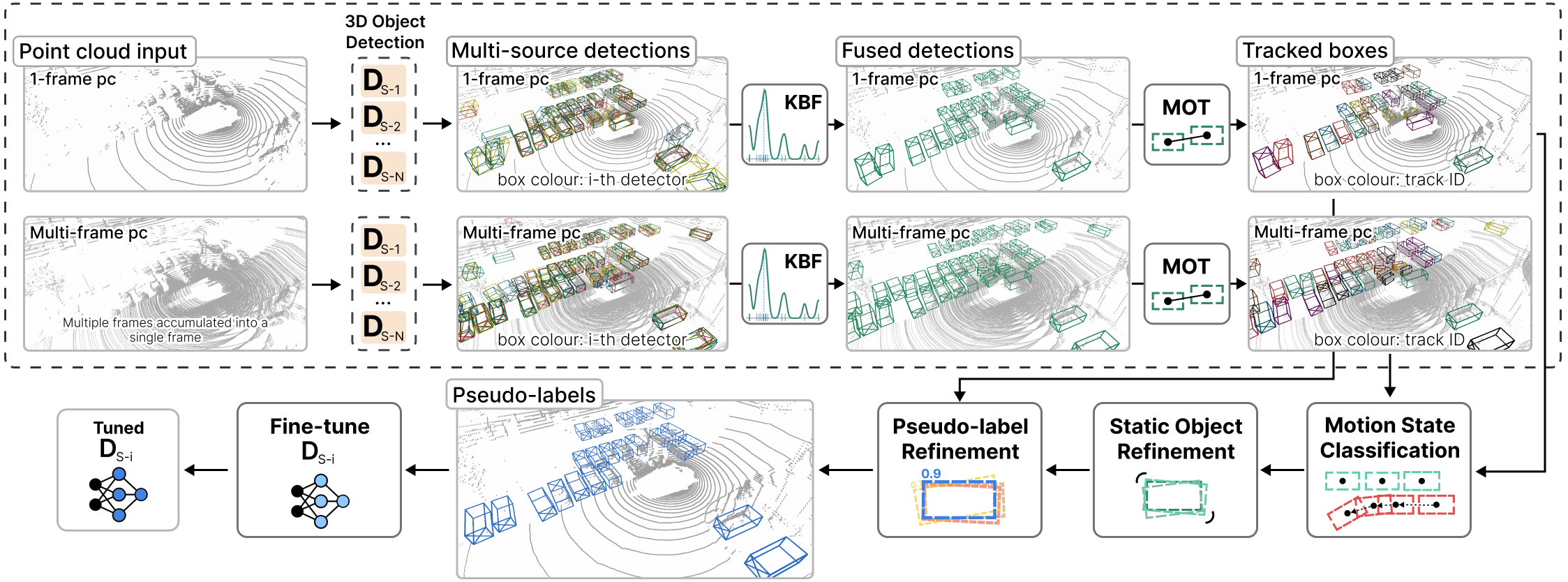}
  \caption{\textbf{The MS3D self-training pipeline.} Given a set of N pre-trained 3D detectors from multiple source domains $\textbf{D}_{\text{S-i}}$ where $\text{i}=1,2,...,\text{N}$, we generate detections for both 1-frame and multi-frame accumulated point clouds, and combine them with our KBF. Thereafter, we feed the sequence frames into a 3D tracker for generating trajectories of both 1-frame and multi-frame boxes. Finally, we refine static object boxes to get our final pseudo-labels for fine-tuning any pre-trained detector $\textbf{D}_{\text{S-i}}$. Visualized in this figure is our MS3D pipeline for Waymo/Lyft $\rightarrow$ nuScenes.}
  \label{fig:pipeline}
  \vspace{-4mm}
\end{figure*}

Our proposed approach is built upon two fundamental observations. Firstly, we recognize that different detectors trained on various datasets possess distinct expertise. For instance, a detector trained on the Waymo dataset may be better suited for a specific novel dataset compared to a detector trained on nuScenes, and vice versa. Secondly, we observe that by utilizing temporal point clouds (i.e., multi-frame accumulation), we can aggregate various perspectives to gain a more comprehensive understanding of static objects' complete geometry, leading to more accurate bounding boxes  \cite{yang20213dman,qi2021offboard,caesar2020nuscenes,chen2022mppnet}.

In real-world scenarios, Multi-Source Domain Adaptation (MSDA) is a practical and effective approach for adaptation to a target domain. The availability of multiple source domains enables us to harness diverse knowledge and improve the overall performance of 3D object detectors on unlabelled datasets. Fortunately, there are many large-scale datasets that can be leveraged for MSDA in 3D object detection, including nuScenes, Waymo, and Lyft, among others. By incorporating the rich information from these sources, we can significantly reduce the domain shift and improve the generalization ability of 3D detectors. 

A naive approach to combining multiple source-trained detectors on a new domain is to select the detection box produced by the most confident detector. However, this method neglects potentially valuable information provided by the other detectors, and it is possible that the most confident detector is not necessarily the most accurate. Instead, we propose KDE Box Fusion (KBF), a more robust box-fusion strategy that utilizes KDE to consider each detector's proposal and enhance pseudo-label quality. We note that KBF can also be applied for detector ensembling in a supervised setting.


Point cloud accumulation has gained popularity as a promising direction for improving single-frame detection through densification. Yang et al. \cite{yang20213dman} demonstrated that concatenating 4 point cloud frames had the optimal performance boost before it started to deteriorate with more frames for the Waymo dataset. However, they observe that for static and slow objects, accumulating more than 4 frames continued to improve detection performance. Qi et al. \cite{qi2021offboard} showed that accumulating 100+ frames of a sequence can be effectively utilized by separating static and dynamic vehicle box refinement. To capitalize on these findings, we explore the use of multi-frame detection in the context of UDA to refine the box localization of static objects. Furthermore, by extending the static boxes of parked vehicles in both directions temporally, we obtain precise pseudo-labels for objects located beyond 50m from the ego-vehicle.


\noindent {\bf Contributions.} Our main contribution is MS3D, a multi-source domain adaptation approach for 3D object detection that leverages multiple source domain experts and temporal information to generate high quality pseudo-labels for self-training. The benefits of our approach are:
\begin{itemize}
\itemsep0em   
  \item {\bf Robust}: By combining the expertise of different detectors trained on various source domains, MS3D is more robust against domain shifts. 
  \item {\bf Versatile and Scalable}: MS3D is easily combined with multiple and various types of pre-trained 3D detectors to further boost pseudo-label quality. 
  \item {\bf Improved Pseudo-labels}: MS3D self-trained models can obtain state-of-the-art UDA performance regardless of the source pre-trained model chosen for fine-tuning. This is because our fused pseudo-labels consistently outperform the best individual source domain detector.
  \item {\bf Speed}: MS3D does not add any processing latency during inference, ensuring it can be used in real-time applications.
  \item {\bf Detection Performance}: MS3D achieves state-of-the-art on all UDA settings presented.

\end{itemize}


\section{Related Work}
\noindent{\bf 3D object detection.} A significant portion of the current research in 3D detection is concentrated on single-frame detection. Within this, many works propose to enhance the feature representation of point clouds, by proposing feature representations that can be classified as voxel-based \cite{yan2018second, lang2019pointpillars}, point-based \cite{shi2019pointrcnn, qi2017pointnet}, or a combination of both \cite{shi2020pv}. Recent works have delved into the use of temporal information, where some works have shown that a simple multi-frame concatenation can already outperform single-frame detection \cite{caesar2020nuscenes, yin2021centernet, yang20213dman}. However, this improvement is limited and the performance deteriorates for longer sequences. To address this challenge, a few studies have focused on developing spatio-temporal feature encoding strategies that fully leverage information from longer sequences \cite{yang20213dman, chen2022mppnet}. Another strategy is to separately encode sequences of dynamic and static object \cite{qi2021offboard} for box refining.

\noindent{\bf Unsupervised Domain Adaptation.} 
Research in UDA for 3D object detection has been predominantly focused on adapting a single detector, trained on a labelled domain (source dataset), to a new domain (target dataset). UDA approaches can be categorised into adversarial methods \cite{qin2019generatively, saito2019strong}, domain-invariant feature representation \cite{yi2021complete}, self-training \cite{zou2018unsupervised}, mean teacher \cite{tarvainen2017meanteacher}, amongst others. More specifically in 3D object detection, prior research showcased successful adaptation of detectors from larger-car datasets like Waymo and nuScenes, to smaller-car datasets like KITTI \cite{wang2020train, yang2021st3d,luo2021mlcnet, tsai2022see, tsai2022viewer}. A popular direction for this topic is to use self-training, where the focus is on generating high quality pseudo-labels for fine-tuning the source-trained detector \cite{yang2021st3d,you2022exploiting,li2022mono3dst}. ST3D \cite{yang2021st3d} employed random object scaling augmentation within a self-training framework for more precise pseudo-labels for adaptation to the KITTI dataset. \cite{wei2022lidardistillation} improved ST3D for adapting from a 64 to 32-beam dataset via their generative method of emulating a pseudo 32-beam lidar.

\noindent {\bf Multi-Source Domain Adaptation.} MSDA is an emerging variant of UDA that builds on UDA techniques while incorporating new techniques that leverage multiple source domains \cite{ahmed2021msda, dong2021confident}. For example, \cite{he2021multi} uses style transfer to reduce the discrepancy of multiple source image styles in a collaborative learning framework. \cite{yao2021multi} proposes to align the feature space of multiple sources through an exponential moving average parameter combination for object detection. Recently, \cite{zhang2023uni3d} extended this to 3D object detection with a focus on learning domain-agnostic features from multiple source datasets to improve detector generalization. To the best of our knowledge, \cite{zhang2023uni3d} is the only attempt at MSDA for 3D object detection. Our approach differs from theirs in that we focus on combining detectors rather than datasets, and we accomplish this without modifying detector architecture or using source domain data (i.e., source-free). Our fusion of proposals from multiple detectors shares similarities with popular ensembling box filtering strategies such as NMS\cite{neubeck2006nms}, Soft-NMS\cite{bodla2017softnms} and Weighted Box Fusion (WBF) \cite{solovyev2021wbf}. In particular, WBF uses confidence scores as weights to determine new box corners in image detection. However, we find that applying WBF to the 3D space is suboptimal compared to our proposed KBF.

\section{Multi-Source 3D}
\subsection{Problem Statement}
In unsupervised MSDA, we have multiple $M$ labelled source domains $S_1,S_2,...,S_M$ and a single unlabelled target domain $T$. In our context, each $i$-th source $S_i=\{(p_{S,i},L_{S,i})\}^{N}_{i}$ is a detector that has been trained on a point cloud dataset $p_{S,i}$ with annotations $L_{S,i}$. This source-trained detector's box predictions on the $k$-th frame of the unlabelled target dataset is denoted as $\{B_{T,i}^{k}\}^{N}_{k}$. Our goal is combine box predictions from each source-trained detector and utilize temporal information to robustly localize and classify objects in 3D for each frame. We assume that the vehicle ego-pose is known in world coordinates for ego-motion compensation. \cref{fig:pipeline} illustrates our proposed pipeline, which we will introduce in the following sections.

\subsection{Fusing multiple sources}
\label{sec:fusing_multiple_source}
\noindent {\bf Initial Pseudo-labels.} In order to enhance the resilience of our model against domain shifts in the unlabelled target domain, we use different types of detectors and source datasets for a greater range of knowledge. For example, with Waymo as our target domain, we employ both nuScenes and Lyft source domains, with two types of detectors, namely SECOND \cite{yan2018second} and CenterPoint \cite{yin2021centerpoint}. We notice that Lyft's lidar scan pattern is similar to that of Waymo, which could give Lyft-trained detectors an edge over nuScenes-trained detectors in certain scenarios. Additionally, different detectors possess varying areas of expertise. For instance, SECOND is adept at estimating object dimensions for axis-aligned objects due to its anchor boxes, whereas CenterPoint performs better with objects that are not axis-aligned \cite{yin2021centerpoint}. 

We take advantage of a detector's full expertise by using two augmentations for Test Time Augmentation (TTA) - random world flip (RWF) across x and y axes, and random world rotation (RWR) in the range $[-\pi, \pi]$. For each source detector we run four permutations: no TTA, RWF, RWR, RWF+RWR, which gives bounding boxes $\{B_{T,i}^{k}\}^{4}_{i}$ for the $i$-th source detector on the $k$-th frame which we then fuse with KBF as follows.

\begin{figure}[t]
  \centering
  \begin{subfigure}{0.49\linewidth}
    \includegraphics[width=0.99\linewidth]{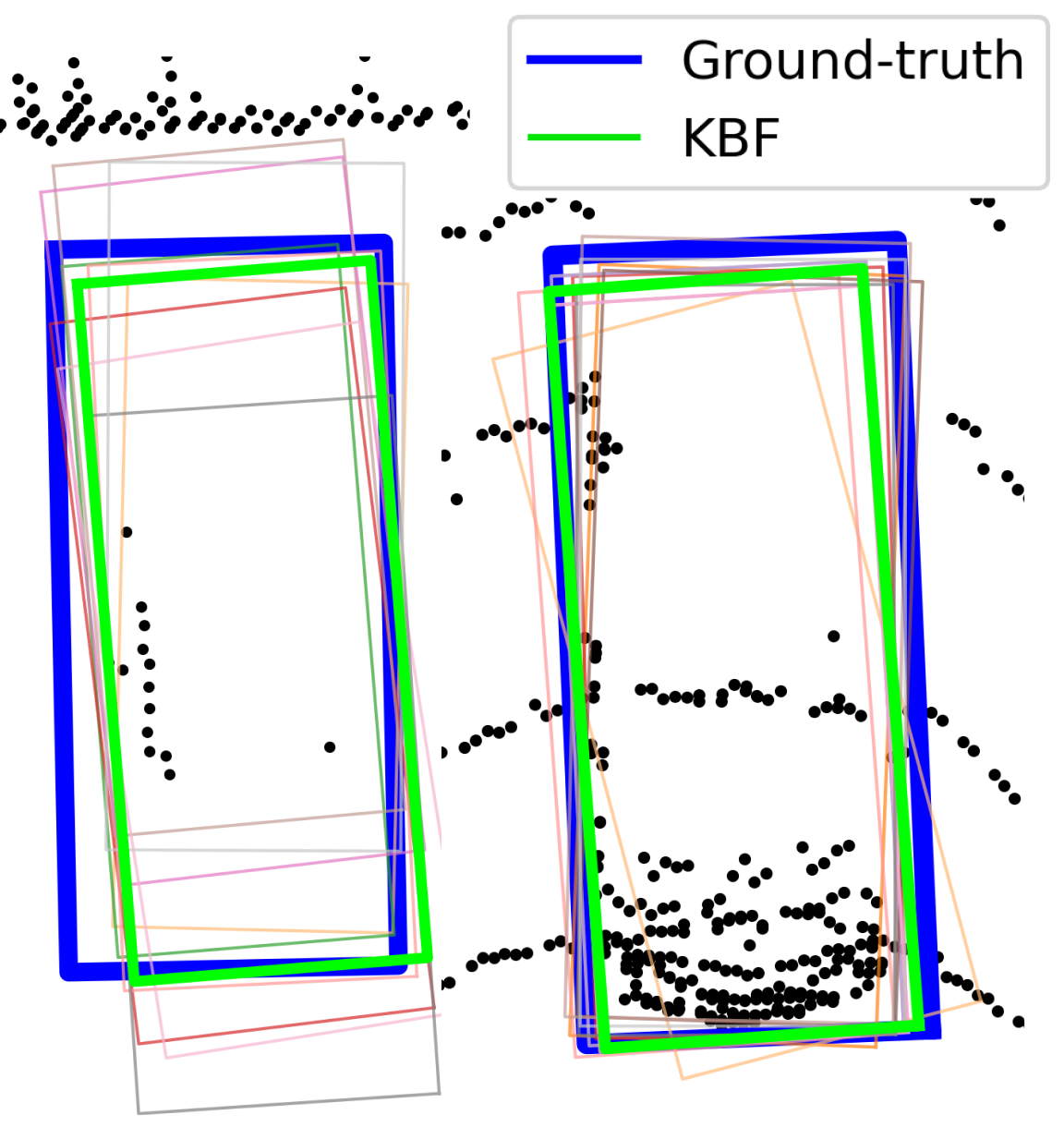}
    \caption{}
    \label{fig:kbf}
  \end{subfigure}
  \hfill
  \begin{subfigure}{0.49\linewidth}
    \includegraphics[width=0.99\linewidth]{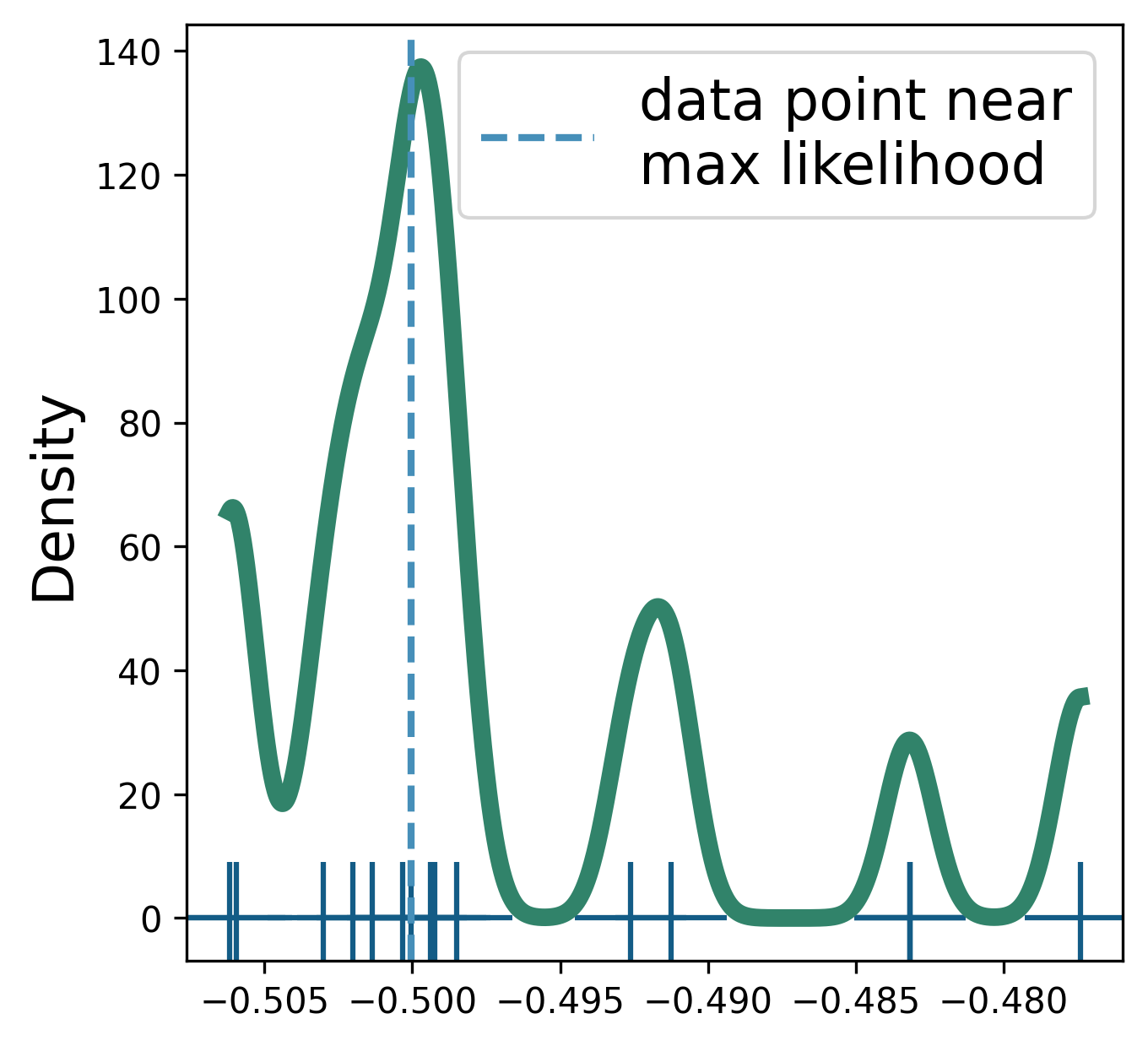}
    \caption{}
    \label{fig:kde}
  \end{subfigure}
  \caption{(a) Fused TTA predictions of 4 source detectors (multi-coloured boxes) with KDE Box Fusion (KBF); (b) Example of how we select the best rotation of multiple boxes with KDE.}
  \label{fig:kbf_kde}
  \vspace{-6mm}
\end{figure}

\noindent {\bf KDE Box Fusion.} The backbone of our KBF method, $\kappa(\cdot)$, is KDE \cite{parzen1962estimation}, which is capable of estimating a probability density function (PDF) using a kernel function shown in \cref{eqn:kde}. KDE places a kernel $K$ at each data point $x_i$ which sums to give the final PDF estimate $\hat{f}(x)$ as shown in \cref{fig:kde}. A weight $\text{w}_i$ for each data point and a kernel bandwidth $h$ can be adjusted for fine-tuning the PDF estimate. 

\begin{equation}
\hat{f}(x) = \frac{1}{h}\sum^{N}_{i=1}\text{w}_iK(\frac{x-x_i}{h})
\label{eqn:kde}
\end{equation}

In 3D object detection, a predicted box is parameterised by its centre $(c_x,c_y,c_z)$, dimensions $(l,w,h)$, heading $\theta$ and a confidence score $s$. Regardless of the localization accuracy, each box carries potentially valuable information. For example, we observe that a box with poor heading estimation may have accurate object dimensions. Therefore to combine the information of every box proposal, we opt to fuse the centre, length, width, height and heading separately. 

To merge boxes, we first create a KD-Tree with the box centroids. We identify matching boxes by querying for sets of boxes that are within a certain radius of each other. We compute KDE separately for the centroid, dimensions and score, then select the peak value of each of their PDF. We take the sine of the heading, $\text{sin}(\theta)$, to ensure rotational continuity before applying KDE. We found empirically that selecting rather than combining box headings produces the best result. Therefore, we choose the box heading with the highest likelihood based on the output PDF. For each box parameter we use a Gaussian kernel and use the predicted box confidence score as weights. Given the set of 1-frame detections from $i$ source detectors, we fuse them with KBF to obtain boxes $B_{\text{1f}}=\kappa(\{B_{i,\text{1f}}\})$ for each frame $k$.

\subsection{Utilizing Temporal Information for Static Vehicles}
\label{sec:using_temporal_information}
\begin{figure}[t]
  \centering
  \includegraphics[width=0.98\linewidth]{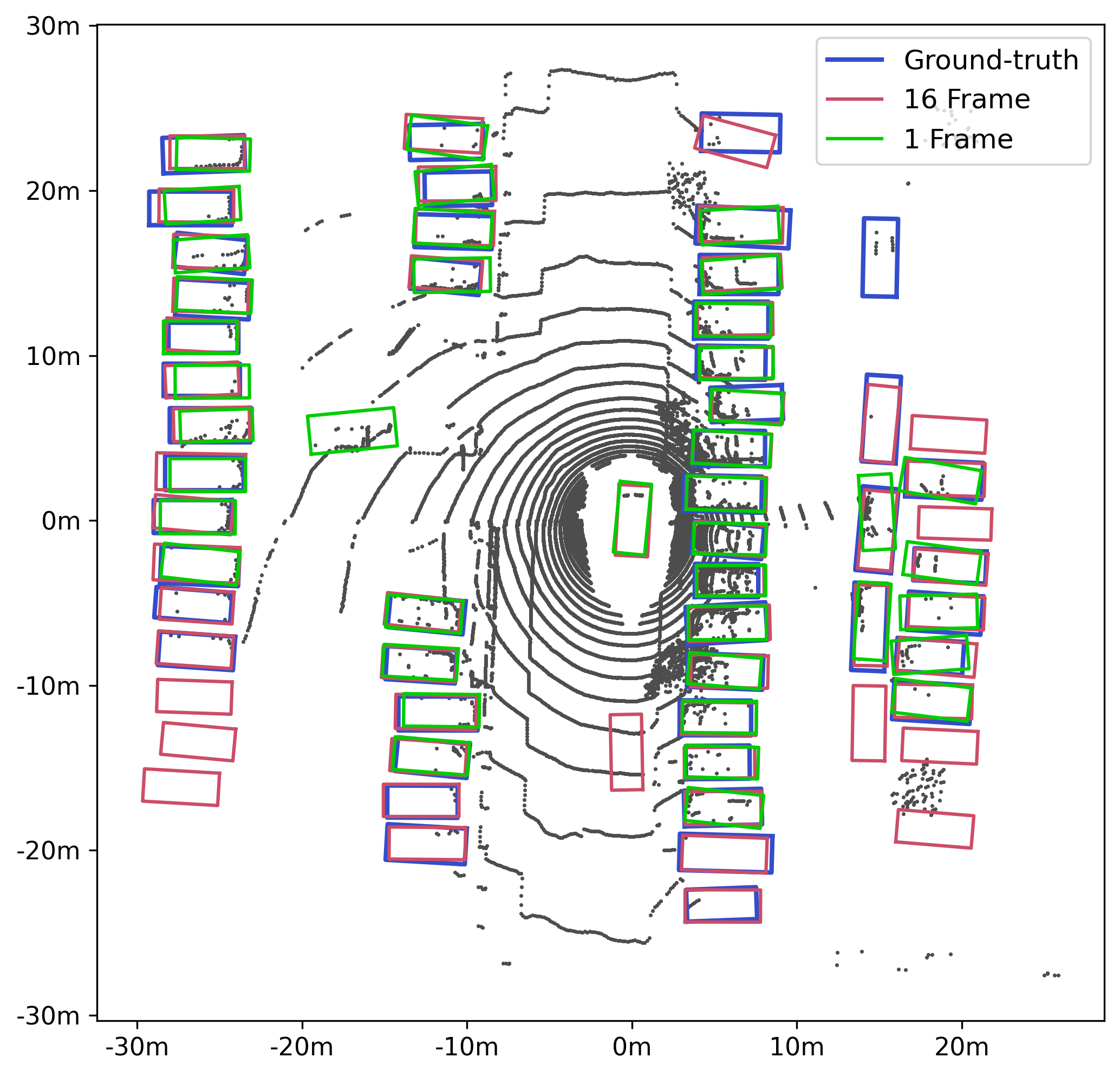}
  \caption{1-frame KBF boxes compared to 16-frame KBF boxes on the nuScenes dataset with detectors trained on Waymo/Lyft as the source domain. Visualized is a 1-frame point cloud.}
  \label{fig:1fvs16f}
  \vspace{-4mm}
\end{figure}

In urban city environments, parked vehicles are common and often are completely static for long periods. Car parks in particular are often present in datasets and provides an abundance of viewpoints for various types of vehicles. With this in mind, we use multi-frame detection to leverage the additional viewpoints provided by accumulated point clouds to better localize static vehicles. For multi-frame detection, the input point cloud $p^k_t$ for frame $k$ is the stacking of $N$ historical point clouds with the point cloud at time $t$ where each historical point cloud has been transformed to the ego-frame at time $t$. With a 10Hz lidar, each point cloud is 0.1s apart, and the accumulated point cloud can be represented as $p^k_t=\{p_t, p_{t-0.1}, p_{t-0.2},...,p_{t-N(0.1)}\}$. By feeding the source detector with $p^k_t$, it can utilize the increased point density to improve the accuracy of box localization, particularly for static objects that may be occluded or located at a distance as shown in \cref{fig:1fvs16f}. In particular, we follow \cite{chen2022mppnet} to use $N=16$ frames for detection for a trade-off between TTA inference time and point cloud density. We use the same TTA permutations and combine box predictions with KBF, $B_{\text{16f}}=\kappa(\{B_{i,\text{16f}}\})$, as in \cref{sec:fusing_multiple_source}.

\subsection{Motion State Classification}
\label{sec:motion_state_classification}
To assign temporally consistent boxes for static vehicles, we associate detection boxes across frames to classify the motion state of the object. To accomplish this, we transform all boxes to world coordinates with known sensor poses. Thereafter, we utilize a tracking-by-detection, Kalman Filter-based tracker \cite{pang2021simpletrack} which is an implementation variant of \cite{Weng2020ab3dmot} to generate box trajectories for both 1-frame, $T_{\text{1f}}$, and 16-frame, $T_{\text{16f}}$, predictions. Following \cite{qi2021offboard}, we classify the tracklets as either static or dynamic based on pre-defined thresholds of begin-to-end distance and box centre variance. 

Nonetheless, this is not sufficient in classifying $T_{\text{16f}}$ due to predictions on the motion trails of dynamic objects in 16-frame detection. We observe that when two dynamic vehicles are driving in the same lane at a similar speed, their motion tails often overlap, leading to a false 'static' classification. A simple approach to replacing static boxes would be to rely solely on the motion classification of $T_{\text{1f}}$ and replace static objects with their corresponding $T_{\text{16f}}$ boxes. However, this approach is suboptimal since 1-frame detection often detects less static objects at a distance than 16-frame detection for certain target datasets as shown in \cref{fig:1fvs16f}, causing us to overlook valuable 16-frame boxes as a result.

To improve the accuracy of $T_{\text{16f}}$ motion classification, we make a key observation: in 3D space, dynamic objects cannot occupy the same space as an object that is static throughout the sequence. This means that the trajectory of a dynamic object will never overlap with that of a parked vehicle. To take advantage of this, we use $T_{\text{1f}}$ to identify all dynamic objects in frame $k$, and project their entire trajectory $T_{\text{1f}}$ into frame $k$. This effectively identifies drivable areas and enables us to filter out wrongly classified static boxes through comparing IOU-matched $T_{\text{1f}}$ and $T^k_{\text{16f}}$ boxes. If either $T_{\text{1f}}$ or $T^k_{\text{16f}}$ is labelled as dynamic, we classify the object as dynamic. While this might lead to more objects falsely classified as dynamic, it prevents harmful extrapolation of vehicles that are only temporarily stationary.

\subsection{Multi-frame Static Object Refinement}
\label{sec:multiframe_static_refinement}
\begin{figure}[t]
  \centering
  \includegraphics[width=0.96\linewidth]{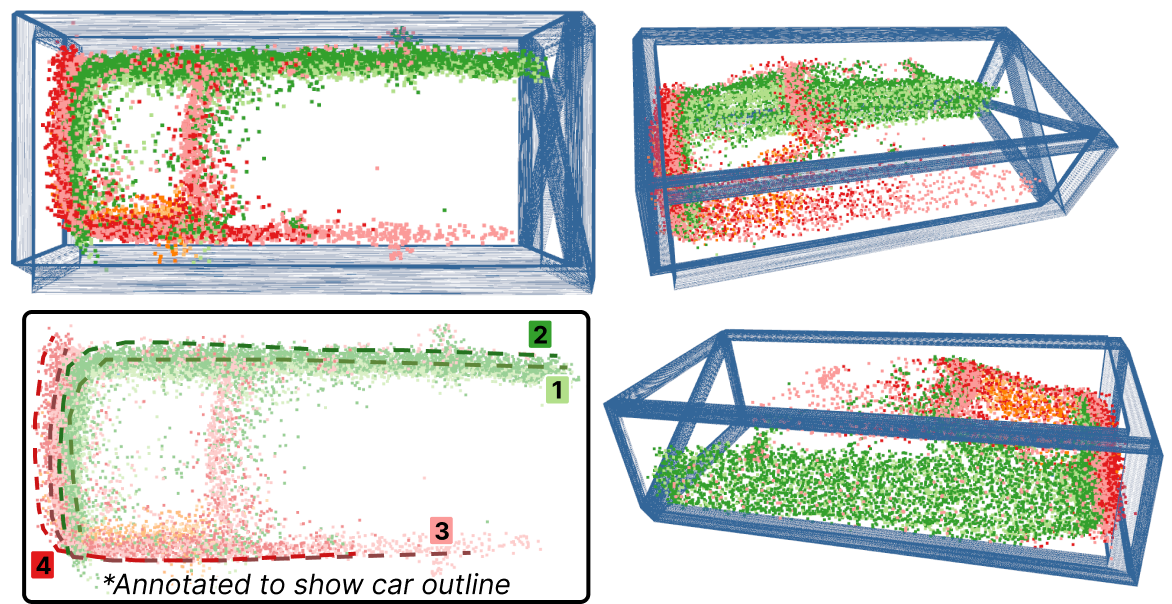}
  \caption{Accumulated frames of a parked car and its assigned ground-truth boxes from all frames of a Waymo sequence (197 frames/boxes total). \textbf{\color{LimeGreen}Light green}, \textbf{\color{OliveGreen}dark green}, \textbf{\color{Lavender}pink}, \textbf{\color{red}red} and \textbf{\color{Orange}orange} represent accumulated points in intervals of 40 frames in order where light green and orange represents frames 0-40 and 160-197 respectively. The bottom left annotation illustrates the registration misalignment, especially between light green (1) and red (4) outlines on the car's rear, potentially caused by ego-pose drift. Best viewed in colour.}
  \label{fig:ego_pose_error}
  \vspace{-5mm}
\end{figure}

We anticipate that a parked car's position in the real world should remain constant. However, we have observed that attempts to assign a single bounding box to a park car for an entire sequence often leads to the exclusion of some car points in certain frames. To investigate this, we accumulated points and ground truth boxes from all frames in a sequence and noticed many parked cars had poor registration across frames. Moreover, the bounding box positions were inconsistent. \cref{fig:ego_pose_error} shows that annotators assign multiple boxes to a parked car over 197 frames due to poor registration, potentially caused by ego-pose drift. Selecting any ground truth box for the car may lead to the partial or complete exclusion of object points. This highlights the fact that relying on a single bounding box to label a parked vehicle throughout a sequence can be unsatisfactory.

This poor registration of parked vehicles was observed in multiple datasets such as nuScenes, Waymo, Lyft and KITTI with some datasets worse than others. To address this, we propose to use KBF on $H$ historical observation frames, $\kappa(\{T_{\text{16f}}\}^k_{k-H})$, to obtain a static box for the $k$-th frame instead of using all frames. This gives us a static box $\{B_{\text{static}}^{k}\}^{N}_{k}$ for each $k$-th frame with a score determined by $\text{max}(\alpha,s_{\text{static}}^{k}))$ where $\alpha$ is a minimum score threshold and $s_{\text{static}}^{k}$ is the fused score of $\{B_{\text{static}}^{k}\}^{N}_{k}$.

To extend the range of pseudo-labels, we assume that static vehicles with sufficiently long tracks will remain static throughout the sequence. Many source-trained detectors can only reliably identify vehicles within a closer range for a new target domain. Therefore, we propagate the first and last KBF boxes of static vehicles through historical and future frames, respectively, under the assumption that they remain static. To reduce the increase in false positives resulting from this propagation, we decay the confidence score of the propagated boxes by $\beta$ for each frame that it is propagated. For certain datasets, we found that updating the pseudo-labels in the self-training process to refine these propagated boxes can give a bump in performance.

\subsection{Pseudo-Label Refinement}
\label{sec:ps_generation}
In the UDA setting, higher point density with 16-frame accumulation does not necessarily lead to improved detection at close ranges compared to 1-frame detections. This is partly because densified point clouds may accentuate differences in scan patterns, and also because static boxes can be susceptible to ego-pose drift, which negatively affects localization accuracy. To address this issue and combine all box proposals, we employ Non-Maximum Suppression (NMS) to allow confident single frame detections to replace 16-frame static boxes. We project the following into the scene and apply NMS for filtering: (1) 1-Frame KBF boxes; (2) $T_{\text{1f}}$ tracked boxes; and (3) Static object boxes, $\{B_{\text{static}}^{k}\}^{N}_{k}$. Finally, we remove any pseudo-labels with less than 1 point inside their box. By including $T_{\text{1f}}$ tracked boxes, we can interpolate missing detections and extrapolate for farther ranges in a similar manner to \cite{you2022exploiting}.

\section{Experiments}
\subsection{Setup}
\label{sec:exp_setup}
\noindent {\bf Datasets.} Many works have addressed the object size domain gap and improved adaptation to the KITTI dataset \cite{yang2021st3d,wang2020train,tsai2022viewer,luo2021mlcnet,you2022exploiting}. However, other factors such as scan pattern, weather and geographical differences have a significant impact on the domain gap. To focus on this, we select datasets with similar object sizes: Waymo \cite{sun2020waymo}, Lyft\cite{woven2019lyft} and nuScenes\cite{caesar2020nuscenes}, summarised in \cref{tab:datasets}, for our experiments. 

\setlength{\tabcolsep}{0.2em} 
\begin{table}
\centering
\begin{tabular}{ccccccc} 
\hline
Dataset  & Lidar Beams     & Size     & Rain & Night & Country  \\ 
\hline
nuScenes & 1$\times$32            & 34,149         & Yes  & Yes    & USA, Singapore \\
Waymo    & 1$\times$64 $+$ 4$\times$200    &  192,484      & Yes  & Yes    &  USA \\
Lyft     & 1$\times$40 or 64 $+$ 2$\times$40 & 18,634       & No   & No    & USA   \\
\hline
\end{tabular}
\caption{Overview of each dataset. Country refers to the country in which the dataset was collected.}
\label{tab:datasets}
\vspace{-4mm}
\end{table}

\noindent{\bf Methods.} We compare MS3D with the following methods: (1) Source-only, which is the direct evaluation of a source-trained detector on the target dataset; (2) SN as a baseline to observe how much of the domain gap is caused by size differences; (3) ST3D, the state-of-the-art self-training approach for 3D object detection; (4) Lidar Distillation, an approach that builds on ST3D for the Waymo to nuScenes setting; (5) GT Fine-tune, the best performance we can achieve when we fine-tune the source-trained detector on the target ground-truth labels. We compare methods (1)-(4) with GT fine-tune as it is a more realistic scenario of what is achievable in the presence of perfect pseudo-labels, as opposed to careful selection of the voxel sizes and anchor box dimensions for an ``Oracle" supervised benchmark \cite{yang2021st3d}. We fine-tune all models with no frame accumulation. We only compare Lidar Distillation for the target-nuscenes setting as it is designed specifically for the high beam to low beam UDA setting. Though we use SECOND-IoU and CenterPoint for generating pseudo-labels, we only focus on the validation results of SECOND-IoU from different source domains to highlight domain discrepancies. 

\noindent {\bf Evaluation Metric.} We follow \cite{yang2021st3d,wang2020train,wei2022lidardistillation} and adopt the KITTI evaluation metric of Average Precision (AP) over 40 recall positions, in Bird's Eye View (BEV) and 3D, denoted as $\text{AP}_{\text{BEV}}$/$\text{AP}_{\text{3D}}$. We assess all models on the ``Vehicle" category. For Lyft and nuScenes, we map ``truck" and ``bus" to the ``car" category. We report the AP at Intersection over Union (IoU) thresholds of 0.7 and 0.5 i.e. a vehicle is considered a true positive detection if the IoU of the predicted and ground truth box is over 0.7 or 0.5. We present results at IoU=0.5 as well to highlight that our method greatly improves the number of correctly detected vehicles.

\begin{table}
\centering
\def\arraystretch{1.1}%
\begin{tabular}{c|c|c|c|c} 
\hline
\multicolumn{5}{c}{Target: nuScenes}                                                                  \\ 
\hline
Method                      & Source                 & Detector    & IoU=0.7       & IoU=0.5        \\ 
\hline
\multirow{4}{*}{Source-Only} & \multirow{2}{*}{Waymo} & SECOND  & 32.91 / 17.24 & 43.32 / 37.58  \\
                             &                        & CenterPoint & 32.10 / 17.77 & 40.83 / 35.77  \\ 
                             & \multirow{2}{*}{Lyft}  & SECOND  & 24.15 / 13.47 & 30.52 / 26.79  \\
                             &                        & CenterPoint & 23.71 / 11.31 & 30.45 / 26.17  \\
                             
\hline
\multirow{2}{*}{SN \cite{wang2020train}}          & Waymo                  & SECOND  & 33.23 / 18.57 & 43.19 / 37.74  \\
                             & Lyft                   & SECOND  & 27.51 / 17.00 & 33.32 / 29.92  \\ 
\hdashline
\multirow{2}{*}{ST3D \cite{yang2021st3d}}        & Waymo                  & SECOND  & 35.92 / 20.19 & 43.03 / 38.99  \\
                             & Lyft                   & SECOND  & 29.88 / 18.37 & 33.18 / 30.67  \\ 
\hdashline
Lidar Dist. \cite{wei2022lidardistillation}                  & Waymo                  & SECOND  & 42.04 / 24.50 & - / -          \\ 
\hdashline
\multirow{2}{*}{MS3D (Ours)} & Waymo                  & SECOND  & \textbf{42.23 / 24.76} & \textbf{52.33 / 46.65}  \\
                             & Lyft                   & SECOND  & \textbf{41.64 / 23.46} & \textbf{51.47 / 45.74}  \\ 
\hline
GT Fine-Tune                 & Waymo                  & SECOND  & 44.39 / 29.46 & 55.61 / 50.83  \\
\hline
\end{tabular}
\caption{{\bf UDA from Waymo/Lyft to nuScenes.} We report $\text{AP}_{\text{BEV}}$/$\text{AP}_{\text{3D}}$. Best results for each source-target pair are highlighted in \textbf{bold}. GT Fine-tune uses GT labels to fine-tune the pre-trained detectors with no frame accumulation.}
\label{tab:target_nusc_results}
\vspace{-5mm}
\end{table}

\noindent {\bf Implementation Details.} We train SECOND-IoU \cite{yan2018second} and CenterPoint \cite{yin2021centerpoint} on all datasets and select the best pre-trained model based on evaluating with the source dataset labels. We re-use the same source pre-trained detectors for adapting to all settings. We argue that this is more realistic to real-world applications as we may not have sufficient labels to evaluate and tune the best model for adaptation to a new target dataset. All our models were trained with the OpenPCDet \cite{openpcdet2020} codebase with tweaked settings for voxel size of $(0.1$m, $0.1$m, $0.15$m$)$, detection range of $[-75,75]$m for x,y axes and $[-2,4]$ for z-axis, and we shift the point cloud frame to the ground plane. Lyft and nuScenes pre-trained models were trained with 5 and 10 frame accumulation respectively, following popular practice \cite{woven2019lyft,caesar2020nuscenes}.

For MS3D, we adopt the settings of ST3D \cite{yang2021st3d} and fine-tune all pre-trained detectors with no frame accumulation, for 30 epochs at a learning rate of $1.5\times 10^{-3}$ with the Adam one-cycle scheduler. We use pseudo-labels with score threshold of $s > 0.6$ as the ground-truth labels for fine-tuning. For fusing proposals with KBF, we consider a vehicle as present if there are more than 4 box proposals with their centroids within 2m of each other. We use $H=16$ historical frames for static object refinement and propagate them with a decay of $\beta=0.95$ with a minimum score $\alpha=0.7$ if they have more than 7 tracked detections. To focus on static vehicles, we sort Waymo and nuScenes scenes by number of cars and select the top 150 and 190 scenes from nuScenes and Waymo respectively for training; scenes with many cars tend to be carparks. For an unlabelled target domain, this can similarly be done by counting the number of fused 1-frame detections per scene, or by intentionally collecting data from scenes with many parked vehicles. Additionally, this suggests that our model can be further improved with a second round of self-training with the remaining scenes which we leave for future work.

\subsection{Results}
\label{sec:results}
As shown in \cref{tab:target_lyft_results,tab:target_nusc_results,tab:target_waymo_results}, our approach outperforms all the other methods regardless of which source domain is used. 
With nuscenes as target domain, \cref{tab:target_nusc_results}, MS3D not only improves detections with accurate box localizations (IoU=0.7) and outperforms existing approaches but also greatly increases the number of correct object proposals (IoU=0.5) by a large margin. It can be seen that whilst SN and ST3D improved box localization with IoU=0.7, they did not greatly increase the number of true positive detections as IoU=0.5 remains relatively unchanged. In contrast, MS3D obtains both a high $\text{AP}_{\text{3D}}$ in IoU=0.7 and IoU=0.5 criterions. In particular, we highlight that for IoU=0.5, MS3D is only $\sim 4\text{AP}$ less than the GT Fine-Tune approach, indicating that we can nearly detect all vehicles in the scene. This is also apparent with Waymo as target, \cref{tab:target_waymo_results}, where MS3D obtains a high 42.88 $\text{AP}_{\text{3D}}$ for IoU=0.7 and $72.34 / 69.30$ $\text{AP}_{\text{BEV}}$/$\text{AP}_{\text{3D}}$ for IoU=0.5 which is  $\sim 3\text{AP}$ less than GT Fine-Tune. With Lyft as the target domain, \cref{tab:target_lyft_results}, MS3D also outperforms other methods in both IoU criterions. From the above results, we have demonstrated that MS3D is able to adapt detectors from low (e.g. nuScenes) to high (e.g. Waymo) beam lidar datasets and vice-versa, unlike ST3D \cite{yang2021st3d} and Lidar Distillation \cite{wei2022lidardistillation}.

\begin{table}
\centering
\def\arraystretch{1.1}%
\begin{tabular}{c|c|c|c|c} 
\hline
\multicolumn{5}{c}{Target: Waymo}                                                                        \\ 
\hline
Method                      & Source                    & Detector    & IoU=0.7       & IoU=0.5        \\ 
\hline
\multirow{4}{*}{Source-Only} & \multirow{2}{*}{Lyft}     & SECOND  & 52.14 / 34.40 & 61.66 / 58.71  \\
                             &                           & CenterPoint & 50.56 / 31.02 & 62.32 / 58.97  \\
                             & \multirow{2}{*}{nuScenes} & SECOND  & 50.30 / 24.70 & 61.40 / 53.73  \\
                             &                           & CenterPoint & 49.39 / 24.33 & 59.52 / 52.64  \\ 
\hline
\multirow{2}{*}{SN \cite{wang2020train}}          & Lyft                      & SECOND  & 53.39 / 39.22 & 62.42 / 59.55  \\
                             & nuScenes                  & SECOND  & 50.69 / 28.86 & 61.38 / 53.90  \\ 
\hdashline
\multirow{2}{*}{ST3D \cite{yang2021st3d}}        & Lyft                      & SECOND  & 56.06 / 39.17 & 65.16 / 62.39  \\
                             & nuScenes                  & SECOND  & 55.67 / 28.83 & 67.19 / 61.63  \\ 
\hdashline
\multirow{2}{*}{MS3D (Ours)} & Lyft                      & SECOND  & \textbf{61.25 / 42.88} & \textbf{72.34 / 69.30}  \\
                             & nuScenes                  & SECOND  & \textbf{61.39 / 42.76} & \textbf{72.47 / 69.45}  \\ 
\hline
GT Fine-Tune                 & nuScenes                  & SECOND  & 66.76 / 52.50 & 75.16 / 72.40  \\
\hline
\end{tabular}
\caption{{\bf UDA from nuScenes/Lyft to Waymo.} We report $\text{AP}_{\text{BEV}}$/$\text{AP}_{\text{3D}}$.}
\label{tab:target_waymo_results}
\vspace{0mm}
\end{table}

\begin{table}
\centering
\def\arraystretch{1.1}%
\begin{tabular}{c|c|c|cc} 
\hline
\multicolumn{5}{c}{Target: Lyft}                                                                                           \\ 
\hline
Method                      & Source                    & Detector    & IoU=0.7                & IoU=0.5                 \\ 
\hline
\multirow{4}{*}{Source-Only} & \multirow{2}{*}{Waymo}    & SECOND  & 49.68 / 38.94          & 56.22 / 54.63           \\
                             &                           & CenterPoint & 43.83 / 32.98          & 49.94 / 48.42           \\
                             & \multirow{2}{*}{nuScenes} & SECOND  & 42.22 / 23.62          & 48.45 / 45.51           \\
                             &                           & CenterPoint & 45.98 / 26.36          & 52.41 / 49.43           \\ 
\hline
\multirow{2}{*}{SN \cite{wang2020train}}          & Waymo                     & SECOND  & 71.61 / 56.13          & 80.65 / 78.52           \\
                             & nuScenes                  & SECOND  & 63.11 / 39.60          & 72.27 / 68.25           \\ 
\hdashline
\multirow{2}{*}{ST3D \cite{yang2021st3d}}        & Waymo                     & SECOND  & 73.86 / 56.33          & 82.90 / 80.81           \\
                             & nuScenes                  & SECOND  & 67.33 / 41.82          & 77.17 / 72.99           \\ 
\hdashline
\multirow{2}{*}{MS3D (Ours)} & Waymo                     & SECOND  & \textbf{76.48 / 61.23} & \textbf{83.36 / 81.49}  \\
                             & nuScenes                  & SECOND  & \textbf{75.02 / 59.01} & \textbf{83.46 / 81.52}  \\ 
\hline
GT Fine-Tune                 & Waymo                     & SECOND  & 81.10 / 66.76          & 91.12 / 88.69           \\
\hline
\end{tabular}
\caption{{\bf UDA from nuScenes/Waymo to Lyft.} We report $\text{AP}_{\text{BEV}}$/$\text{AP}_{\text{3D}}$. GT Fine-tune uses Lyft labels to fine-tune the pre-trained detector with no frame accumulation.}
\label{tab:target_lyft_results}
\vspace{-5mm}
\end{table}


For each target domain, we show the detection performance from two different source domains. We demonstrate that with MS3D, regardless of the pre-trained detector's source dataset, the detector can be fine-tuned to a similar state-of-the-art performance. In contrast, ST3D's choice of pre-trained detector's source dataset can have a significant impact on performance, with a 10.34 $\text{AP}_{\text{3D}}$ difference at IoU=0.7 between selecting nuScenes or Lyft. MS3D's robustness is especially valuable in real-world scenarios where acquiring sufficient labelled data for evaluation can be costly.

\begin{figure}[t]
  \centering
  \includegraphics[width=0.96\linewidth]{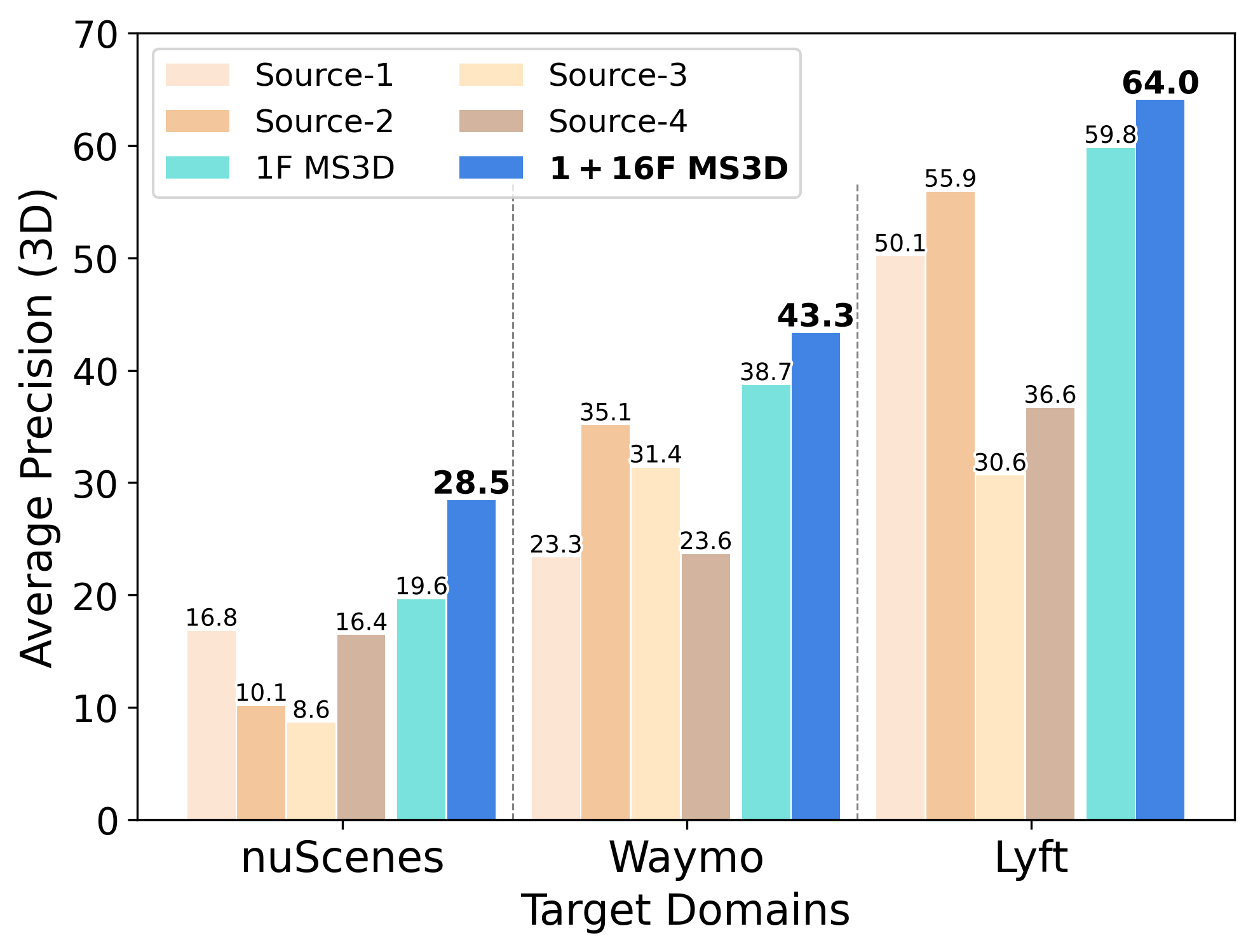}
  \caption{Evaluation of MS3D's pseudo-labels on different target domains. 1F MS3D refers to using only KBF on 1-frame detections from the 4 source domains. 1+16F MS3D refers to our final pseudo-labels after refining static vehicles.}
  \label{fig:ps_quality}
  \vspace{-5mm}
\end{figure}

\subsection{Ablation}
\label{sec:ablation}
\noindent {\bf Pseudo-label Quality.} In \cref{fig:ps_quality}, we show that our pseudo-labels consistently improve the detections of every source-trained detector on the target domain. When we combine all 4 sources with KBF, we can already surpass the best source-trained detector's $\text{AP}_{\text{3D}}$. By adding 16-frame detections for refining of static vehicles, we can further boost the detection by an impressive amount for all target domains. For example with nuScenes as the target domain, MS3D overcomes the detection range limitations of a sparse-beam lidar with our multi-frame static refinement, leading to a 8.84 $\text{AP}_{\text{3D}}$ increase from 1F MS3D to 1F+16F MS3D.

\noindent {\bf KDE Box Fusion.} We compare KBF with NMS \cite{neubeck2006nms} and two variants of Weighted Box Fusion \cite{solovyev2021wbf} in \cref{tab:ablation_kbf}. WBF-C is the official 3D implementation using opposite 3D box corners, and WBF-P is weighted average of box params $(c_{xyz},l,w,h,\theta,s)$. We show that KBF outperforms other methods at all ranges, with the 50-80m range having the most improvement over existing methods. With KDE, setting a high bandwidth is similar to taking the average of the data points. Being able to tune the bandwidth for different parameters allows us to better capture the nuances of the box fusion process. For example, a common error in box prediction is the heading orientation being incorrect by an error of $\pi$. Setting a high bandwidth for rotation estimation would therefore be detrimental as we may select an in-between orientation instead of one that corrects the orientation by $\pi$.

\noindent {\bf Static Object Refinement.} In \cref{tab:ablation_rkde}, we show that using 16-frame detections for static objects ($H=0$) gives a slight improvement over 1F MS3D (i.e., using KBF for 1-frame detections). Applying KBF on $H > 0$ historical boxes of a static object can further improve the localization of the box. As explained in \cref{sec:multiframe_static_refinement}, using KBF on all historical boxes is inadequate. This is shown in the lower performance of ``Single Box" compared to using $H=16$ boxes. Quantifying the extent of this difference is challenging because comparing $H=16$ and ``Single box" to a GT label may only differ by a few cm, which can still meet the IoU=0.7 threshold. Errors in point cloud alignment over long sequences becomes problematic when there are sparse number of points on the car, as a few cm difference could result in the bounding box excluding the car's points entirely. Therefore, based on qualitative and quantitative assessment, we chose to use $H=16$ boxes for estimating the bounding box of a parked vehicle. 

\begin{table}
\centering
\def\arraystretch{1.1}%
\begin{tabular}{c|c|ccc} 
\hline
\multicolumn{1}{l|}{}  & IoU=0.7       & \multicolumn{3}{c}{RANGE (L2/APH)}  \\
Method                 & $\text{AP}_{\text{BEV}}$/$\text{AP}_{\text{3D}}$        & {[}0,30) & {[}30,50) & {[}50,+inf)  \\ 
\hline
nuScenes / CenterPoint & 49.49 / 23.33 & 58.80    & 17.56     & 1.89         \\
nuScenes / SECOND      & 50.74 / 23.64 & 60.07    & 15.56     & 1.61         \\
Lyft / CenterPoint     & 49.26 / 31.35 & 61.53    & 30.97     & 9.88         \\
Lyft / SECOND          & 50.83 / 35.09 & 46.98    & 19.11     & 6.93         \\ 
\hline
NMS \cite{neubeck2006nms}                 & \textbf{57.93} / 34.88 & 57.13    & 24.01     & 7.59         \\
WBF-C \cite{solovyev2021wbf}                    & 53.27 / 36.55 & 67.17    & 32.74     & 8.33         \\
WBF-P \cite{solovyev2021wbf}                 & 54.44 / 36.75 & 67.43    & 33.32     & 8.92         \\
KBF (Ours)                    & 57.22 / \textbf{38.67} & \textbf{68.89}    & \textbf{33.93}     & \textbf{10.40}       \\
\hline
\end{tabular}
\caption{Ablation study of box fusion methods for multiple detector predictions on Waymo as target domain.}
\label{tab:ablation_kbf}
\vspace{0mm}
\end{table}

\begin{table}
\centering
\def\arraystretch{1.1}%
\begin{tabular}{c|c|ccc} 
\hline
\multicolumn{1}{l|}{}  & IoU=0.7                & \multicolumn{3}{c}{RANGE (L2/APH)}                \\
Method                 & $\text{AP}_{\text{BEV}}$/$\text{AP}_{\text{3D}}$                  & {[}0,30)       & {[}30,50)      & {[}50,+inf)     \\ 
\hline
nuScenes / CenterPoint & 49.49 / 23.33          & 58.80          & 17.56          & 1.89            \\
nuScenes / SECOND      & 50.74 / 23.64          & 60.07          & 15.56          & 1.61            \\
Lyft / CenterPoint     & 49.26 / 31.35          & 61.53          & 30.97          & 9.88            \\
Lyft / SECOND          & 50.83 / 35.09          & 46.98          & 19.11          & 6.93            \\ 
\hline
1F MS3D	& 57.22 / 38.67	& 68.89	& 33.93	& 10.4 \\
\hline
$H=0$                    & 59.19 / 38.73          & 69.49          & 34.47          & 10.63           \\
$H=4$                    & 62.06 / 40.89          & 71.74          & 38.83          & 14.11           \\
$H=16$                   & \textbf{62.32 / 43.33} & \textbf{72.23} & \textbf{41.72} & 17.63           \\
Single box             & 62.32 / 43.12          & 72.02          & 41.30          & \textbf{17.88}  \\
\hline
\end{tabular}
\caption{Ablation study of our static box labelling using a rolling window of $H$ frames for KBF on Waymo as target domain. For $H=0$ we replace static boxes with their 16-frame detections. For ``Single box", we use KBF of all observed frames.}
\label{tab:ablation_rkde}
\vspace{-5mm}
\end{table}

\subsection{Discussion}
In our experimentation with various source to target domain pairs, we observed that while 16-frame accumulation increased point density, it did not necessarily help detectors localize well even when the entire car shape is visible. For example, when testing nuScenes/Waymo detectors on KITTI, we found that for a fully visible, 16-frame accumulated car at 5m away from the ego vehicle, detectors consistently predicted a box that was 0.5-1m longer. This was also observed when using a KITTI detector on nuScenes/Waymo where it would predict a smaller box that did not encapsulate the entire object. This is likely due to the scan pattern domain gap as 16-frame accumulated point clouds appear quite distinct depending on the lidar used. To address this gap when the target dataset statistics deviate significantly from the source dataset, we recommend incorporating Random Object Scaling \cite{yang2021st3d} or SN \cite{wang2020train} in the self-training process. 

\section{Conclusion}
In this paper, we introduce MS3D, an innovative self-training pipeline that utilizes multiple source domains and temporal information to adapt to new target domains.  Leveraging the availability of multiple labelled datasets in real-world applications, we demonstrate that combining detectors trained on these sources can effectively auto-label an unlabelled target dataset and fine-tune the source detector to boost performance. MS3D achieves state-of-the-art results on all tested UDA settings, with our proposed KBF method consistently outperforming the best individual source detector in terms of improving pseudo-label quality. While the optimal combination of detectors for an unlabelled target domain remains an open question, MS3D lays a strong foundation to integrate new and existing detector architectures and facilitates low-cost deployment in real-world scenarios.

\bibliography{references}
\bibliographystyle{IEEEtran}

\end{document}